# ARTS: An Open-Source, Reproducible Three-Arm Platform for Autonomy Research in Minimally Invasive Surgery

Jiawei Ge[1,2,†*], Chunwoo Kim[3,†], Antony Goldenberg[1,2,†], Douglas McDonough[1], Justin Liao[1], Bo Ning[4], Sylvia Ladstaetter[4], Katya Van Anderlecht[4], Hao Ding[2], Laura Connolly[1], Ethan Kilmer[1], Francis Fan[1], Shouyue Hu[1,2], Tyler Salvador[4], Brendan Burkhart[1], Pit Henrich[1], Richard Cha[4], Tito Porras[2], Mathias Unberath[1,2], Kevin Cleary[4], and Axel Krieger[1,2]

*Abstract*— **Research on surgical autonomy needs open platforms that any laboratory can build, yet the most widely shared platform depends on donated clinical hardware, and open alternatives on industrial arms each lack part of the surgical chain. We present the Autonomous Robotic Translational System (ARTS), an open-source three-arm platform for autonomy research in minimally invasive surgery, built entirely from purchasable parts for about USD 137,000. Two UR5e arms hold wristed da Vinci X/Xi instruments, a UR5 arm holds a stereo endoscope with two FLIR cameras, the surgeon console provides haptic handles, footswitches and a stereo viewer, and a hardware stop independent of the software lets a supervisor freeze all three arms in place. The ROS 2 software pins every dependency, installs from the same scripts on bare metal or in containers, and records stereo images with hardware-measured timing. On the real arms, ARTS commands each arm at its controller's full rate (e.g., 500 Hz on the UR5e), the two cameras expose within 12 $\mu$s of each other, hand motion reaches the arm in 58 ms, and the endoscope image reaches the monitor in 68 ms. The ISO 9283 pose repeatability is 0.14-0.25 mm, comparable to a da Vinci Research Kit (dVRK) Classic arm on the same protocol, and an optical tracker measured a mean remote-center-of-motion error of 0.34-0.37 mm. Three expert and three novice operators performed the Fundamentals of Laparoscopic Surgery (FLS) peg transfer through a laparoscopic trainer box, and one of the experts sutured on a suture pad. The hardware designs, firmware and software are available at https://github.com/MRE-Lab-UMD/arts.**

## I. INTRODUCTION

Robot-assisted surgery (RAS) gives surgeons dexterity, magnified vision and better ergonomics, but outcomes still depend heavily on individual skill, and teleoperation eases rather than eliminates the fatigue that leads to errors [1]. Autonomy holds the potential to improve accuracy and consistency: a model-based autonomous robot first performed laparoscopic intestinal anastomosis in vivo under human supervision [2]. More recently, learning-based methods have moved soft-tissue autonomy further: hierarchical imitation learning completed the clipping and cutting phase of cholecystectomy ex vivo [3], and laparoscopic task autonomy has been shown in vivo [4]. These methods depend on data, and surgical robot learning now faces a data bottleneck: datasets are small, come from one robot at one institution, and are rarely shared, because most research runs on devices engineered for clinical use, each with its own hardware, software and interfaces [5]. This calls for a research platform that drives clinical instruments, gives direct access to the endoscopic images, and records synchronized data, and that other labs can rebuild.

Raven-II was among the first surgical robots built as a shared research platform, with custom arms, instruments and electronics later sold by a spin-off company [6], [7]. The da Vinci Research Kit (dVRK) is the most widely shared open platform [8], and imitation-learning results such as SRT-H were obtained on it [3], but it runs on retired da Vinci hardware [9] that a lab cannot purchase; access depends on donated clinical systems. Open platforms built on industrial arms remove that dependency, and each covers part of the chain. A Modular and Open Platform for Surgical Robotics Research (MOPS) drives wristed da Vinci instruments on UR arms with a software remote center of motion (RCM) [10]. An open laparoscopic platform pairs UR arms with a non-wristed laparoscopic instrument and a Karl Storz stereo endoscope, and deploys learned policies [11]. DASH brings da Vinci instruments to a Unitree G1 humanoid and is designed to fit serial-link arms such as the Franka Panda [12]. None of them, however, combines a robotic stereo endoscope, wristed instruments with the energy delivery that is routine in clinical robotic surgery, and hardware-synchronized stereo recording in one system that a lab can purchase and install (Table I).

We present the Autonomous Robotic Translational System (ARTS), an open-source three-arm research platform for minimally invasive surgery (MIS) built entirely from commercially available parts (Figure 1). Two UR5e arms drive da Vinci X/Xi instruments, and a UR5 holds a stereo endoscope whose two optical channels are imaged by machine-vision FLIR cameras with open drivers, so the raw images and their exposure times reach the software directly rather

† These authors contributed equally to this work.
* Corresponding author: Jiawei Ge. jge9@jh.edu

This work was supported by the Advanced Research Projects Agency for Health (ARPA-H) under grant number AY1AX000023, and the National Science Foundation (NSF/FRR CAREER) under grant number 2144348.

[1]J. Ge, A. Goldenberg, D. McDonough, J. Liao, L. Connolly, E. Kilmer, F. Fan, S. Hu, B. Burkhart, P. Henrich, M. Unberath, and A. Krieger are with the Laboratory for Computational Sensing and Robotics (LCSR), Johns Hopkins University, Baltimore, MD 21218, USA

[2]J. Ge, A. Goldenberg, H. Ding, S. Hu, T. Porras, M. Unberath, and A. Krieger are with Inner Logic, Baltimore, MD 21211, USA

[3]C. Kim is with the Korea Institute of Science and Technology (KIST), Seoul 02792, Republic of Korea

[4]B. Ning, S. Ladstaetter, K. Van Anderlecht, T. Salvador, R. Cha, and K. Cleary are with the Sheikh Zayed Institute for Pediatric Surgical Innovation, Children's National Hospital, Washington, DC 20010, USA

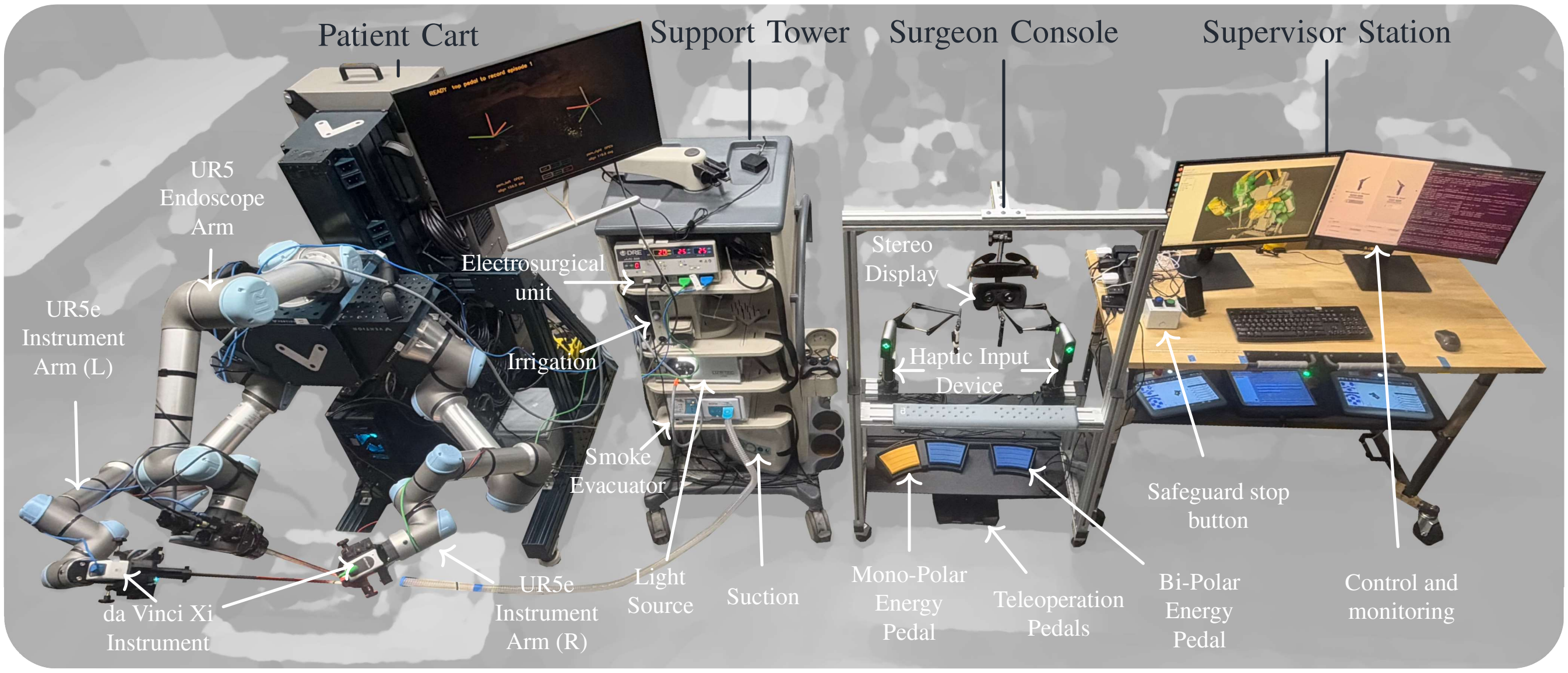

Fig. 1. Overview of the ARTS platform. The patient cart carries two UR5e instrument arms with da Vinci X/Xi instruments and a UR5 stereo endoscope arm. The support tower houses the electrosurgical unit and auxiliary equipment. The surgeon console provides haptic input, a stereo display, and teleoperation and energy pedals. The supervisor station provides system control, monitoring and the safeguard stop button.

TABLE I

OPEN SURGICAL RESEARCH PLATFORMS. ARMS: SOURCE OF THE INSTRUMENT ARMS. ENDO.: ROBOTIC ENDOSCOPE ARM. WRISTED: WRISTED DA VINCI INSTRUMENTS. SYNC.: HARDWARE-SYNCHRONIZED STEREO CAPTURE. – : NOT REPORTED.

| Platform | Arms | Endo. | Wristed | Energy | Sync. | License |
|---|---|---|---|---|---|---|
| Raven-II [6] | custom | – | × | – | – | open |
| dVRK [7], [9] | retired clinical | ✓ | ✓ | – | × | open* |
| MOPS [10] | UR | × | ✓ | – | – | stated† |
| Rodriguez *et al.* [11] | UR, Franka | ✓ | × | – | × | unreleased |
| DASH [12] | G1, Franka | × | ✓ | × | × | MIT, CC BY |
| **ARTS** | UR | ✓ | ✓ | ✓ | ✓ | Apache 2.0 |

*Part of the dVRK-Si arm firmware is closed source. †Stated in the paper; no license file in the public repositories.

than as the video output of a clinical camera system. A haptic surgeon console, a safeguard stop that holds all three arms at once, and a recording pipeline complete the system, and reproduces the software on bare metal or in containers.

## A. *Contributions*

First, we introduce the complete system, with four stations (patient cart, surgeon console, support tower and supervisor station) and a priced bill of materials (BOM) of USD 137,320, and the control hierarchy that lets teleoperation, learned policies and scripted motion drive the arms through one interface with the RCM as a hard constraint. Second, we make the platform reproducible: every part is commercially available and priced in the BOM, the designs of our custom hardware are released, and every software dependency is pinned. Bare-metal and container installations run the same scripts, and the recorded data carry hardware-measured stereo timing (left-right skew $2.7 \pm 3.2\,\mu s$) and device clocks for offline alignment. Third, we evaluate the platform on the real arms with devices independent of the robot's own sensors: an optical tracker for the ISO 9283 pose repeatability (0.14 mm to 0.25 mm) and the physical RCM error (0.34 mm to 0.37 mm), and external cameras for the hand-to-arm (58 ms) and camera-to-monitor (68 ms) latencies.

# II. SYSTEM DESIGN

## A. *Stations and Bill of Materials*

ARTS is arranged as four stations named after their counterparts in the da Vinci system and the dVRK (Figure 1). The *patient cart* is a pedestal (Community Design 524807, Vention, Montreal, Canada) that holds two 6-degree of freedom (DoF) arms (UR5e, Universal Robots, Odense, Denmark) as patient-side manipulators (PSMs), hung at 45° on either side, and one 6-DoF arm (UR5 CB3, Universal Robots) as the endoscopic camera manipulator (ECM) on top. Each PSM carries an instrument adapter that drives a da Vinci X/Xi instrument (Intuitive Surgical, Sunnyvale, CA) through its input discs (Section II-D). The ECM carries a 12 mm stereo endoscope (da Vinci Si, Intuitive Surgical) whose two optical channels are imaged by two machine-vision cameras (Section II-C). The cart also holds the main computer and a safety relay that holds all three arms on command (Section II-E). The *surgeon console* carries two haptic devices (Inverse3 with VerseGrip handle, Haply Robotics, Montreal, Canada) fitted with custom finger grippers, a stereo head-mounted display (HMD) (G3 Max, GOOVIS, NED Optics, Shenzhen, China), a three-pedal teleoperation footswitch (Savant Elite2, Kinesis, Bothell, WA), and monopolar and bipolar energy footswitches (BV-1253B and BV-1254B, Bovie Medical, Clearwater, FL). The *support tower* is a used laparoscopic tower with a monitor, a laparoscopic light source, a smoke evacuator, irrigation and suction pumps, and an electrosurgical unit (ESU) (ASG-300, DRE Medical, Louisville,

KY) whose pedals pass through a custom energy interposer (Section II-D). The *supervisor station* is where the operator starts and monitors the system, and it carries the stop button.

Every part can be purchased from the released BOM, whose priced total is USD 137,320. The three arms account for 66 % of it (costed at a typical quoted price), and the custom components (instrument adapters, stereo coupling, energy interposer and camera synchronization box) for less than 4 %. Clinical equipment is costed used, as a research lab would purchase it, and the UR5 CB3, which is no longer sold new, is costed as a UR5e.

### B. Software Architecture and Installation

The control software runs on a single computer with Ubuntu 24.04 and a real-time (RT) Linux kernel, and is built on Robot Operating System 2 (ROS 2) [13] Jazzy and its standard control framework, ros2_control. Following the ROS 2 convention, the software is organized as workspaces stacked on an underlay: the ROS 2 installation is the underlay, pinned third-party drivers form the first overlays, the ARTS packages the next, and task-specific applications sit on top, so a new application never modifies the layers below it. Each arm runs its own controller manager at the rate of its controller interface (500 Hz for the PSMs and 125 Hz for the ECM), with RT priority on an isolated CPU core. Code, firmware, hardware designs and the BOM are released under the Apache 2.0 license at `https://github.com/MRE-Lab-UMD/arts`.

### C. Stereo Vision and Recording

The two cameras (Blackfly S BFS-U3-63S4C, Teledyne FLIR, Wilsonville, OR) are triggered by a custom synchronization board that also captures each camera's exposure strobe on one 80 MHz counter. The left-right skew of every frame is therefore measured in hardware, with 12.5 ns resolution, rather than estimated from software timestamps. The cameras stream raw Bayer images at $1536 \times 1024$ pixels and 30 Hz with each frame's device timestamp, and two rosbag2 recorders write them and the robot state to separate MCAP files without rewriting any timestamp, so images are joined to robot data offline through the camera clock.

### D. Instruments and Energy

Laparoscopic tasks need wristed instruments, and dissection and hemostasis need energy, so ARTS uses da Vinci X/Xi instruments, including monopolar and bipolar energy instruments. Among open instrument interfaces for industrial arms, such as those of MOPS [10] and DASH [12], we use the instrument module and energy interposer of [14], because they drive these instruments, energy instruments included, and let the surgeon's pedals or software switch the electrosurgical generator. Two of their properties matter at the system level. First, the instrument's wrist and jaw joints are part of the arm's command interface, so one tool-tip target sets the arm, the wrist and the jaw together (Section III). Second, energy activations are published with their source and recorded with the robot and image streams, so energy use becomes part of the data.

### E. Safety

The stop button at the supervisor station is wired through the safety relay to the safeguard inputs of all three arm controllers. Pressing it brings every arm to a controlled stop with its motors powered (a Category 2 stop), so the arms hold their pose without sagging, and motion resumes only after a separate reset. Each arm's teach-pendant emergency stop still cuts its power, and the ESU is stopped at its own power switch. In software, a collision monitor wraps the arm links, instrument adapters, proximal instrument shafts, endoscope and pedestal in capsules and spheres, and holds all arms when any two come within 20 mm.

## III. CONTROL AND TELEOPERATION

### A. Control Hierarchy

Every arm is driven by one position-commanded hierarchy. A *reference source*, the teleoperation follower, a learned policy or a goal mover, publishes the tool-tip target. The *task layer* computes joint rates by resolved-rate control, the target velocity fed forward plus a gain on the pose error, subject to the RCM and the joint limits as hard constraints of a quadratic program (QP) (Section III-C). Kinematics come from Pinocchio [15], the QP is solved by ProxQP [16], and the rates are integrated to joint positions. Ruckig [17] then bounds each joint's velocity, acceleration and jerk, and the result is sent to the arm controller every cycle. On the PSMs the command is the arm controller's joint-velocity interface (`speedj`), which we chose over its position interface (`servoj`) because it followed the master with less lag in an A/B comparison. The ECM uses `servoj`, because under the RCM its velocity interface oscillated. Since teleoperation and autonomy differ only in the reference source, the action recorded for learning is the same in both: the absolute tool-tip target.

### B. Teleoperation

Each haptic device reports its handle position at 500 Hz, and its VerseGrip handle reports orientation and buttons on a slower wireless stream. For the jaw, a 3D-printed grip on the handle rotates a magnet over the Hall-effect sensor under one of the handle's buttons, and the sensor reading is mapped to the jaw opening. The follower maps handle motion, scaled down four times, and handle orientation to the instrument tip in the endoscope view, and feeds the handle velocity forward to the task layer. The console follows da Vinci conventions. The left pedal is a clutch that is enabled only while the HMD's head-presence sensor detects the surgeon, and while the head is away both handles are held in place. The middle pedal hands both masters to the endoscope arm: the motion common to both hands moves the endoscope, turning the line between the hands rolls it about its optical axis, and a haptic bar holds the distance between the hands. A tool arm re-engages only when the handle orientation is within $10^\circ$ of the instrument's, so the instrument does not jump. A pedal panel on both displays shows the pedal and energy states.

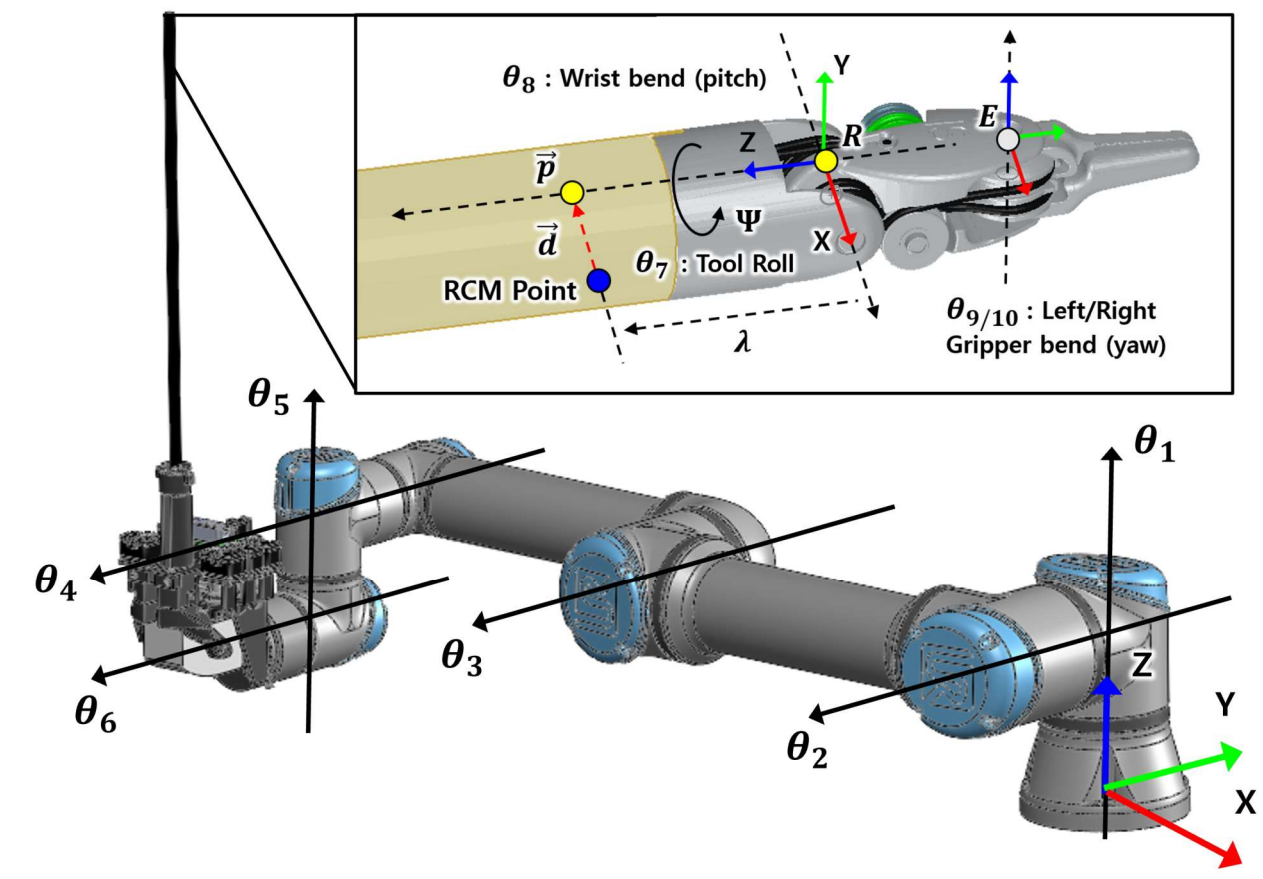

Fig. 2. Joints and coordinate systems of the robot.

### C. Remote Center of Motion Control

Each PSM has ten DoFs: six from the UR5e and four from the instrument (roll, wrist pitch and the two jaws). The controller uses nine of them, because the two jaws enter only through their mean, the grasper rotation $\theta_9 = (j_L + j_R)/2$; the opening $\alpha = j_L - j_R$ does not move the frame at the base of the grasper and is commanded directly. The nine joints serve six task rows, the pose of the grasper frame $B$, and three constraint rows: two keep the shaft on the RCM point, and one stops the UR5e from rolling the shaft, so that all rotation about the shaft comes from the instrument's roll joint and the arm's elbow and shoulder stay clear of the neighboring arms.

Figure 2 shows the instrument arm with its joints and frames. The pose of $B$ is given by the product-of-exponentials forward kinematics ${}^O_BT = e^{\hat{s}_1\theta_1}\cdots e^{\hat{s}_9\theta_9}\,{}^O_BT_0$, where $\hat{s}_i$ is the twist of joint $i$ and ${}^O_BT_0$ is the pose of $B$ at the zero configuration. Joints 1–6 are the UR5e's, joint 7 is the instrument roll, joint 8 the wrist pitch and joint 9 the grasper rotation.

*1) RCM Constraint:* The RCM frame $R$ sits at the distal end of the shaft, before joints 7–9, with its $z$ axis along the shaft (Figure 2). The shaft point $p$ that coincides with the RCM point lies at the distance $\lambda$ from $R$ along the shaft, and it may move only along the shaft. Its velocity in $R$ is ${}^R\vec{v}_p = \vec{v}_R + \vec{\omega}_R \times \lambda\vec{e}_z$, where $\vec{v}_R$ and $\vec{\omega}_R$ are the linear and angular body velocities of $R$ and $\vec{e}_z = [0,0,1]^T$. Requiring its $x$ and $y$ components to cancel the measured offset $\vec{d}$ of the shaft from the RCM point gives two rows,

$$\vec{e}_x^{\,T} J_v \dot{\vec{\theta}} + \lambda\, \vec{e}_y^{\,T} J_\omega \dot{\vec{\theta}} = -k_r\, d_x, \quad (1)$$

$$\vec{e}_y^{\,T} J_v \dot{\vec{\theta}} - \lambda\, \vec{e}_x^{\,T} J_\omega \dot{\vec{\theta}} = -k_r\, d_y, \quad (2)$$

where $J_v$ and $J_\omega$ map the joint rates $\dot{\vec{\theta}}$ to $\vec{v}_R$ and $\vec{\omega}_R$, and $d_x$, $d_y$ are the components of $\vec{d}$ in $R$. With a zero right-hand side, as in the kinematic derivation, an offset left by tracking error would persist; the feedback term $-k_r\vec{d}$ makes it decay instead. The insertion depth $\lambda$ is computed from the measured geometry every cycle and bounded below by 5 mm, so the instrument cannot be withdrawn into the cannula. The third row stops the UR5e from rolling the shaft,

$$\vec{e}_z^{\,T} J_\omega \dot{\vec{\theta}} = -k_\psi\, \psi, \quad (3)$$

where $\psi$ is the shaft roll produced by the UR5e since engagement. Because $R$ precedes the instrument roll, eq. (3) involves joints 1–6 only, and joint 7, whose axis coincides with the shaft, produces every roll. Without this row we observed frequent self-collisions of the UR5e as it rolled the shaft with its own joints.

*2) Quadratic Programming Problem:* The task is the desired pose ${}^O_BT_d$ with the velocity feedforward $\vec{\xi}_{ff}$ from the reference source. Each control cycle solves

$$\min_{\dot{\vec{\theta}}} \left\| W\left(J_B\dot{\vec{\theta}} - (\vec{\xi}_{ff} + K\vec{e})\right)\right\|^2 + \mu \left\|\dot{\vec{\theta}} - k_p(\vec{\theta}_0 - \vec{\theta})\right\|^2 \quad (4)$$

subject to eqs. (1) to (3), the insertion bound, and the joint velocity and position limits. Here $J_B$ is the body Jacobian of $B$, $\vec{e}$ the pose error in the body frame, $K = \mathrm{diag}(k_\omega I_3, k_v I_3)$, $W$ a task weight, and the second term pulls the joints toward the posture $\vec{\theta}_0$ recorded at engagement. With the RCM enabled, the nine joints meet nine conditions, so the solution is unique, and the posture term matters only near singularities and active limits. The joint rates are integrated, bounded in velocity, acceleration and jerk by Ruckig, and then projected back onto eqs. (1) and (2) in closed form, because Ruckig limits each joint independently and its lag would otherwise move the shaft off the RCM point. The controller runs at 500 Hz over all nine joints and sends the instrument joint commands to the instrument module every cycle. The ECM uses the same controller without a wrist: its six joints serve the two RCM rows and four task rows, the endoscope tip position and its roll about the shaft.

## IV. EXPERIMENTS

### A. Experimental Design

All experiments ran on the system of Sections II and III with the real arms.

**Real-time performance:** Kernel scheduling latency was measured with cyclictest on the isolated cores under combined CPU and GPU load for 300 s. The 500 Hz control loop of one PSM was then recorded for 10 min (314142 cycles) of scripted motion; this run used the earlier inverse-kinematics controller without the RCM.

**Stereo capture and recording:** Both cameras were triggered at 30 Hz and recorded for 30.5 min with the surgeon display running, and every frame's left-right skew was taken from the synchronization board.

**Latency:** For teleoperation, a 120-frame-per-second camera filmed the haptic handle and the PSM's adapter in the same frame while the operator moved the handle back and forth (two takes). For video, the endoscope filmed a monitor that showed a Gray-code clock beside the live endoscope view, so each frame contains the clock as it is and as it was one trip through the video path earlier (5 min).

**Pose repeatability:** Following ISO 9283:1998 [18], a ProGrasp forceps on one PSM held a four-sphere marker tracked by an optical tracker (fusionTrack 500, Atracsys, Puidoux, Switzerland), and the arm visited five taught poses 30 times each; each pose was measured once the tracked marker was static. On ARTS, the RCM was off and no cannula was used, and the adapter sets the free shaft length of the instrument to that of the dVRK PSM; the dVRK Classic ran the same protocol, tracker and marker with its cannula mounted, as in its standard configuration.

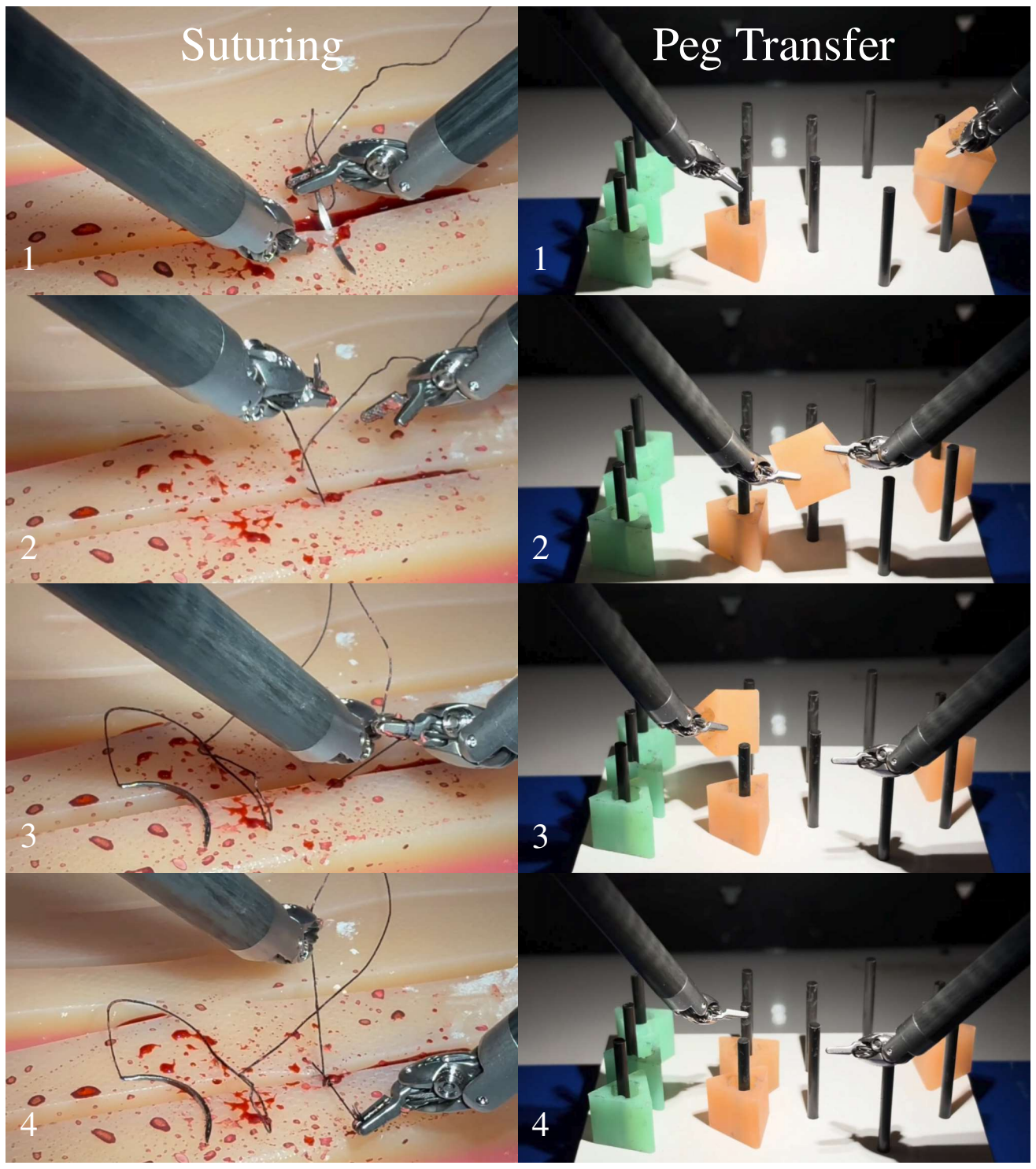

Fig. 3. Representative suturing (left) and peg-transfer (right) sequences performed with ARTS. The key frames progress chronologically from top to bottom. Images were captured with an external camera.

**Physical RCM error:** With the RCM on and the point placed 100 mm behind the wrist pitch axis, the tracker followed a marker on the instrument shaft during three 5 min trials of randomized pivoting (up to 15°) and roll (up to 10°).

**Safeguard stop:** The stop button was pressed 10 times at each of four tool speeds (10 mm s$^{-1}$, 20 mm s$^{-1}$ and 100 mm s$^{-1}$) while a PSM moved, and the stop was read from the arm controller's own stop report.

**User tasks:** Three experts (more than two years of surgical robot teleoperation) and three novices (none) each performed one recorded trial of the FLS peg transfer after 5 min of practice: six triangles were lifted with one needle driver, passed to the other in mid-air and placed on the opposite pegs, with the RCM on at the ports of a laparoscopic trainer box (Szabo-Berci-Sackier, Karl Storz, Tuttlingen, Germany) and a 300 s limit. One novice trial was excluded after a haptic handle stopped sending data, and was repeated with another novice. One of the experts also performed bimanual suturing on a suture pad with two needle drivers, as a qualitative demonstration (Figure 3).

TABLE II

MEASURED PERFORMANCE OF ARTS ON THE REAL ARMS. MEAN ± STANDARD DEVIATION UNLESS NOTED; $N$ IN THE TEXT. RP: ISO 9283 POSE REPEATABILITY; L−R: LEFT MINUS RIGHT.

| Measurement | Result | Target |
|---|---|---|
| Worst-case scheduling latency | 19 µs | <200 µs |
| Control-loop jitter (500 Hz, 10 min) | 3.8 % of period | <10 % |
| Frames per sync pulse (30.5 min) | 100 % | ≥99 % |
| Stereo skew, L−R | 2.7 ± 3.2 µs | <1 ms |
| Hand-to-arm latency (two takes) | 58 ms, 54 ms | ≤100 ms |
| Camera-to-monitor latency | 68 ms | ≤100 ms |
| Pose repeatability RP, per pose | 0.14 mm to 0.25 mm | – |
| dVRK Classic, same protocol | 0.09 mm to 1.24 mm | – |
| Physical RCM error, mean per trial | 0.34 mm to 0.37 mm | ≤1 mm |
| Kinematic RCM deviation, median | 0.026 mm to 0.027 mm | ≤0.1 mm |
| in peg transfer, per arm and trial | 0.022 mm to 0.034 mm | ≤0.1 mm |
| Stopping distance at 20 mm s$^{-1}$ | 1.46 ± 0.01 mm | – |

### B. Evaluation Metrics

**Timing:** the worst-case scheduling latency; the loop period's standard deviation as a fraction of the period (jitter) and the count of missed cycles. **Recording:** recorded frames per synchronization pulse, and the left-right skew as the difference of the two exposure strobes on the board's counter. **Latency:** for teleoperation, the lag that maximizes the cross-correlation of the handle and adapter tracks; for video, the difference of the two clock readings in each frame. **Repeatability:** for each pose, $RP = \bar{l} + 3S_l$, where $l$ is the distance of each attained position from the barycenter of the 30 attained positions, $\bar{l}$ its mean and $S_l$ its standard deviation [18]. **RCM error:** a pivot point is fitted to all tracked shaft poses by least squares, as in DASH [12], and the pivot error is the distance of the tracked shaft point from it, per sample; its drift is the slope of that distance over time. We also report the kinematic deviation, the distance from the commanded RCM point to the shaft line computed from the measured joints, as defined for the open laparoscopic platform [11]. **Stop:** the tool's stopping distance and the time until it is within 1 mm of its final pose, with the drift during the hold and the motion at reset. **Peg transfer:** the time between the operator's two presses of the record pedal, the triangles transferred and dropped, and the kinematic RCM deviation over the trial.

## V. RESULTS

### A. Real-Time Performance

Under combined CPU and GPU load, the worst-case scheduling latency on the isolated cores was 19 µs, 0.9 % of the 2 ms control period (Table II). The PSM control loop ran at 499.99 Hz with a period of 2000 ± 77 µs, a jitter of 3.8 %, and no missed cycles after start-up.

### B. Stereo Capture and Recording

Over 30.5 min, each camera recorded 54827 frames for 54827 synchronization pulses, with no gap in pulse indices, image timestamps or device-clock records. The left-right skew was 2.7 ± 3.2 µs (worst case 11 µs), three orders of magnitude below the 33 ms frame period.

### C. Latency

The arm followed the haptic handle with a lag of 58 ms and 54 ms in the two usable takes. The camera-to-monitor video latency was 68 ms on a 60 Hz monitor; the HMD path was not measured.

### D. Pose Repeatability

RP ranged from 0.14 mm to 0.25 mm across the five poses, the pooled distance to the barycenter was $0.078 \pm 0.043$ mm (maximum 0.21 mm), and the orientation repeatability was $0.10^\circ$ to $0.45^\circ$ about each axis. On the same protocol, the dVRK Classic PSM reached an RP of 0.09 mm to 0.11 mm at three poses and of 0.35 mm and 1.24 mm at the other two, where a single cycle deviated by 0.55 mm and 2.08 mm; pooled over all visits, the distance to the barycenter was $0.063 \pm 0.172$ mm, against $0.078 \pm 0.043$ mm for ARTS.

### E. Physical RCM Error

The pivot error was $0.34 \pm 0.17$ mm, $0.35 \pm 0.18$ mm and $0.37 \pm 0.18$ mm in the three trials (maximum 1.09 mm), and the fitted pivot drifted by less than 0.01 mm $\text{min}^{-1}$.

### F. Safeguard Stop

The stopping distance grew nearly in proportion to speed, from $0.72 \pm 0.01$ mm at 10 mm $\text{s}^{-1}$ to $7.76 \pm 0.06$ mm at 100 mm $\text{s}^{-1}$, and the held tool drifted at most 0.014 mm.

### G. User Tasks

All three experts completed the peg transfer, in 93.2 s to 144.8 s (mean 121.5 s), without dropping a triangle. Two of the three novices completed it, in 162.1 s and 219.1 s, and the third had transferred five of six triangles at the 300 s limit, with one triangle dropped in view. No triangle was lost from view, and the collision monitor never stopped an arm. Through the trainer-box ports, the median kinematic RCM deviation was 0.022 mm to 0.034 mm per arm and trial (maximum 0.252 mm). The suturing demonstration is shown in Figure 3.

## VI. Discussion

The teleoperation latency is dominated by the path to the arm, not by sensing or video. The recorded data reproduce the camera's 58 ms from the handle message to the measured joint motion, and changing only the arm's command interface from `servoj` to `speedj` reduced it from 84 ms to 58 ms, as the delay from a joint command to the measured joint motion fell from 34 ms to 40 ms to 4 ms to 9 ms. DASH reports $20.3 \pm 15.7$ ms on the same kind of measurement [12]; our figure includes the UR5e controller's motion response with the instrument adapter as payload.

The kinematic RCM deviation (median 0.026 mm) is about 13 times smaller than the physical pivot error (0.34 mm to 0.37 mm), because the kinematic value assumes the arm's joint readings and kinematic model are exact, whereas the tracker likely also sees kinematic calibration error, compliance of the adapter and shaft, and its own noise. Both values are needed: the kinematic deviation compares directly with the open laparoscopic platform (median 0.035 mm to 0.079 mm across phantom, ex vivo and in vivo trials) [11], and the physical error with DASH ($3.37 \pm 2.44$ mm and $2.64 \pm 1.54$ mm in two trials) [12]. Our trials were in free air without a cannula, so tissue and trocar forces are not yet included.

Sub-millimeter repeatability matters for learning from demonstrations: a policy trained on recorded tool-tip targets can only be as consistent as the arm that replays them. On the same protocol, the UR5e-based arm is less tight than the dVRK Classic at most poses but showed no outlying cycle, so its repeatability is of the same order as that of the arm most surgical learning results were obtained on. During the peg transfer, the kinematic RCM deviation stayed at the level of the scripted trials while operators moved freely, so the constraint holds under teleoperation through real ports.

Several limitations remain. First, the instrument zero comes from automatic homing followed by a visual touch-up; it repeats for a given instrument but differs by a few degrees between instruments, a confound for learning across instruments, and a per-type alignment jig is future work. Second, the Hall-effect grip reading changes only about every 90 ms in steps of $4.7^\circ$ of jaw, which limits fast grasping; future work will raise its update rate with the designers of the grip add-on and the handle. Third, the head-presence sensor reports the surgeon leaving with a perceptible delay that we have not measured, during which the instrument can still move. Finally, ARTS is designed to be reproduced, but it has not yet been rebuilt or installed by an independent group; a second installation at a partner site is planned and will test the BOM, the documentation and the install path.

## VII. Conclusion

We presented ARTS, an open-source three-arm platform for autonomy research in MIS that is built entirely from purchasable parts for about USD 137,000. Its software installs from the same scripts on bare metal or in containers, its recordings carry hardware-measured stereo timing (skew $2.7 \pm 3.2$ μs), and its arms reach an ISO 9283 repeatability of 0.14 mm to 0.25 mm and a physical RCM error of 0.34 mm to 0.37 mm under teleoperation latencies below 70 ms. Three expert and three novice operators performed the FLS peg transfer on it through a laparoscopic trainer box, and one of the experts sutured on a suture pad. We release the platform so that other groups can build the same system and collect comparable data for surgical robot learning.

## Acknowledgment

We thank Anton Deguet for discussions that drew on his expertise and experience in developing the da Vinci Research Kit and dVRK-Si, which informed the design of ARTS.